\documentclass[conference]{IEEEtran}
\IEEEoverridecommandlockouts
\usepackage{graphicx}
\usepackage{pdflscape}  % Rotates the entire page in the PDF
\usepackage{cite}  
\usepackage{graphicx}
\usepackage{media9}
\usepackage{amsmath,amssymb,amsfonts}
\usepackage{textcomp}
\usepackage{xcolor}
\usepackage{tabularx}
\usepackage{array}
\usepackage{url}
\usepackage{caption} 
\usepackage{geometry}
\usepackage{caption} 
\usepackage{float}
\usepackage{stfloats}
\usepackage{hyperref}
\usepackage{rotating}  % Provides rotation commands

\let\origfigure\figure
\let\endorigfigure\endfigure
\renewenvironment{figure}[1][H]%
  {\origfigure[H]}{\endorigfigure}

\def\BibTeX{{\rm B\kern-.05em{\sc i\kern-.025em b}\kern-.08em
    T\kern-.1667em\lower.7ex\hbox{E}\kern-.125emX}}

\begin{document}

\title{A Novel Zero-Touch, Zero-Trust, AI/ML Enablement Framework for  IoT Network Security}

\author{%
Sushil Shakya, Robert Abbas, Sasa Maric
 \\ 
sushil.shakya@live.vu.edu.au , robert.abbas@vu.edu.au, s.maric@unsw.edu.au
\\
Victoria University, Sydney, Australia
\\
University of New South Wales, Sydney Australia
}
\maketitle

\begin{abstract}

The IoT facilitates a connected, intelligent, and sustainable society; therefore, it is imperative to protect the IoT ecosystem.
The IoT-based 5G and 6G will leverage the use of machine learning and artificial intelligence (ML/AI) more to pave the way for autonomous and collaborative secure IoT networks. Zero-touch, zero-trust IoT security with AI and machine learning (ML) enablement frameworks offers a powerful approach to securing the expanding landscape of Internet of Things (IoT) devices. 
This paper presents a novel framework based on the integration of Zero Trust, Zero Touch, and AI/ML powered for the detection, mitigation, and  prevention of DDoS attacks in modern IoT ecosystems. The focus will be on the new integrated framework by establishing zero trust for all IoT traffic, fixed and mobile 5G/6G IoT network traffic, and data security (quarantine-zero touch and dynamic policy enforcement). We perform a comparative analysis of five machine learning models, namely, XGBoost, Random Forest, K-Nearest Neighbors, Stochastic Gradient Descent, and Naïve Bayes, by comparing these models based on accuracy, precision, recall, F1-score, and ROC-AUC. Results show that the best performance in detecting and mitigating different DDoS vectors comes from the ensemble-based approaches. 

By incorporating network slicing, micro-segmentation, continuous authentication, and resilient 5G/6G strategies, the framework offers robust, scalable security against increasingly sophisticated ransomware-based DDoS attacks.
Zero-touch, zero-trust IoT security with AI/ML enablement is the paradigm of a robust cybersecurity strategy in the age of the 5G/6G-based Internet of Things and Industry 4.0 and 5.0. By integrating these technologies, organizations can effectively secure their IoT environments, protect sensitive data, and maintain business continuity in the face of evolving cyber threats

\end{abstract}

\vspace{\baselineskip}

\section{Introduction}
\label{sec:introduction}

The IoT is revolutionizing industries by connecting tens of billions of devices across healthcare, transport,  smart cities, smart industries, smart mining, smart agriculture, and more. The Internet of Things (IoT) is bringing an influx of difficult-to-secure gadgets to enterprise networks. With the rapid increase in the number of IoT devices, efficiency and automation have increased, but so have the risks in IoT systems. One of the most tangible looming threats-attacks is DDoS, which causes a disruption of critical services and resource exhaustion by using botnets made of compromised devices. The combination of DDoS with ransomware tactics increased the risks by encrypting sensitive data and asking for ransom payments. \vspace{\baselineskip}

The various features brought in by 5G/6G networks further complicate the security of IoT with network slicing, low latency, and edge computing. While these diverse advances will enable diverse and high-performance applications, they will also enlarge the attack surface, for which traditional perimeter-based defenses are no longer effective. Such complex and highly distributed networks demand scalable, adaptive, and intelligent solutions to secure them. \vspace{\baselineskip}

We thereby provide an integrated framework: \textbf{Zero-Touch provisioning}, \textbf{Zero-Trust security}, and \textbf{AI/ML-based threat detection}. While Zero-Touch will automate the onboarding process for a secure device, Zero-Trust will proceed with the process of authentication in order to get rid of implicit trust. AI/ML algorithms allow for real-time anomaly detection and adaptation against evolving threats. Scalable and robust, the proposed solution can mitigate current IoT security risks in 5G/6G environments.

\section{Literature Review}
\label{sec:lit_review}

\subsection{IoT Security}
In the IoT context, the efficiency of traditional security approaches like perimeter defenses and signature-based IDS are relatively falling into inefficiency due to the dynamic nature of IoT ecosystems, which are heterogeneous by nature. Most of them are not even able to mitigate modern threats that include DDoS, which depends on the high number of IoT devices with vulnerabilities to perform service disruption along with resource exhaustion \cite{mirkovic2004taxonomy, zargar2013survey}. The rapid proliferation of IoT devices has expanded the attack surface, necessitating more robust and adaptive security measures \cite{xu2020survey, Cisco2022IoTPrediction}.
\vspace{\baselineskip}

\subsection{ML-Enabled Security}
The potential of ML in strengthening IoT security includes conducting real-time anomaly detection, predictive analytics, and adapting responses against ever-evolving threats. Among these models, ensemble models, especially XGBoost and Random Forest, show the greatest efficiency in identifying deviations in network traffic patterns that could hint at malicious activity \cite{dhaliwal2018, Dai2024XGBoostReview}. While effective, ML-based solutions are still plagued by problems of computational overhead, algorithmic bias, and explainability gaps, making them ineffective and very difficult to deploy at scale in IoT environments \cite{sommer2010closed, Gadepalli2021Explainable}.
\vspace{\baselineskip}

\subsection{Zero-Trust Security and Network Sclicing}
Network slicing is a technique that creates multiple virtual networks on a single physical network. Network slicing enables virtualized networks to operate on the same physical network infrastructure. The basic idea of network slicing is to “slice” the original architecture into multiple logical and independent networks. These sliced networks can then be configured to effectively meet various application needs and service requirements.
 Network slicing and Micro-segmentation can not only make up for the shortcomings of firewalls but when implemented network-wide, can remove the need for a firewall altogether. Zero-Trust security ensures continuous authentication and authorization of devices and communications to remove implicit trust within the network boundary. It follows the principle "never trust, always verify," which drastically reduces the chances of lateral movement on the part of the attacker \cite{KindervagZeroTrust, Chang2023ZeroTrust}. Coupled with real-time network traffic analysis and dynamic policy enforcement, this makes it highly suitable for the protection of IoT ecosystems-especially in the context of 5G/6G \cite{Bland2022ZeroTrust5G, Freed2023ZTA}.
\vspace{\baselineskip}

\subsection{Zero-Touch Provisioning}
Zero-Touch provisioning automates the supply chain for onboarding IoT devices in a secured manner with minimum human intervention. This reduces the chances of configuration errors, ensuring that each device is authenticated, securely booted, and checked against the specified policies before joining the network \cite{Cisco2022IoTPrediction, Freed2023ZTA}. This is particularly important for the Zero-Touch scaling of security as IoT networks continue to increase in number and complexity.
\vspace{\baselineskip}

\subsection{Methodology and Dataset Details}
\label{sec:methodology}

For this work, we used a labeled dataset of IoT network traffic that had been anonymized so as to be able to apply machine learning analysis. The dataset contains typical features of network traffic, such as packet size and rate, session duration, protocol type, and labels indicating whether the traffic was part of a DDoS attack or normal activity.

\vspace{\baselineskip}
To prepare the data for analysis, we conducted several preprocessing steps:
\begin{itemize}
    \item \textbf{Data Cleaning}: Removed incomplete or corrupt records that could skew results.
    \vspace{\baselineskip}
    \item \textbf{Feature Selection}: Selected relevant features based on their correlation with DDoS attack detection.
    \vspace{\baselineskip}
    \item \textbf{Normalization}: Standardized feature values to ensure consistency in scale, using Min-Max scaling. For each feature \(x\), the normalized value \(x'\) is calculated as:
    \begin{equation}
        x' = \frac{x - \min(x)}{\max(x) - \min(x)}
    \end{equation}
\end{itemize}

\subsection{Selected ML Models}
We evaluated five machine learning models based on their suitability for classification tasks within cybersecurity contexts[21].

\begin{figure}[H]
\centering
\includegraphics[width=0.48\textwidth]{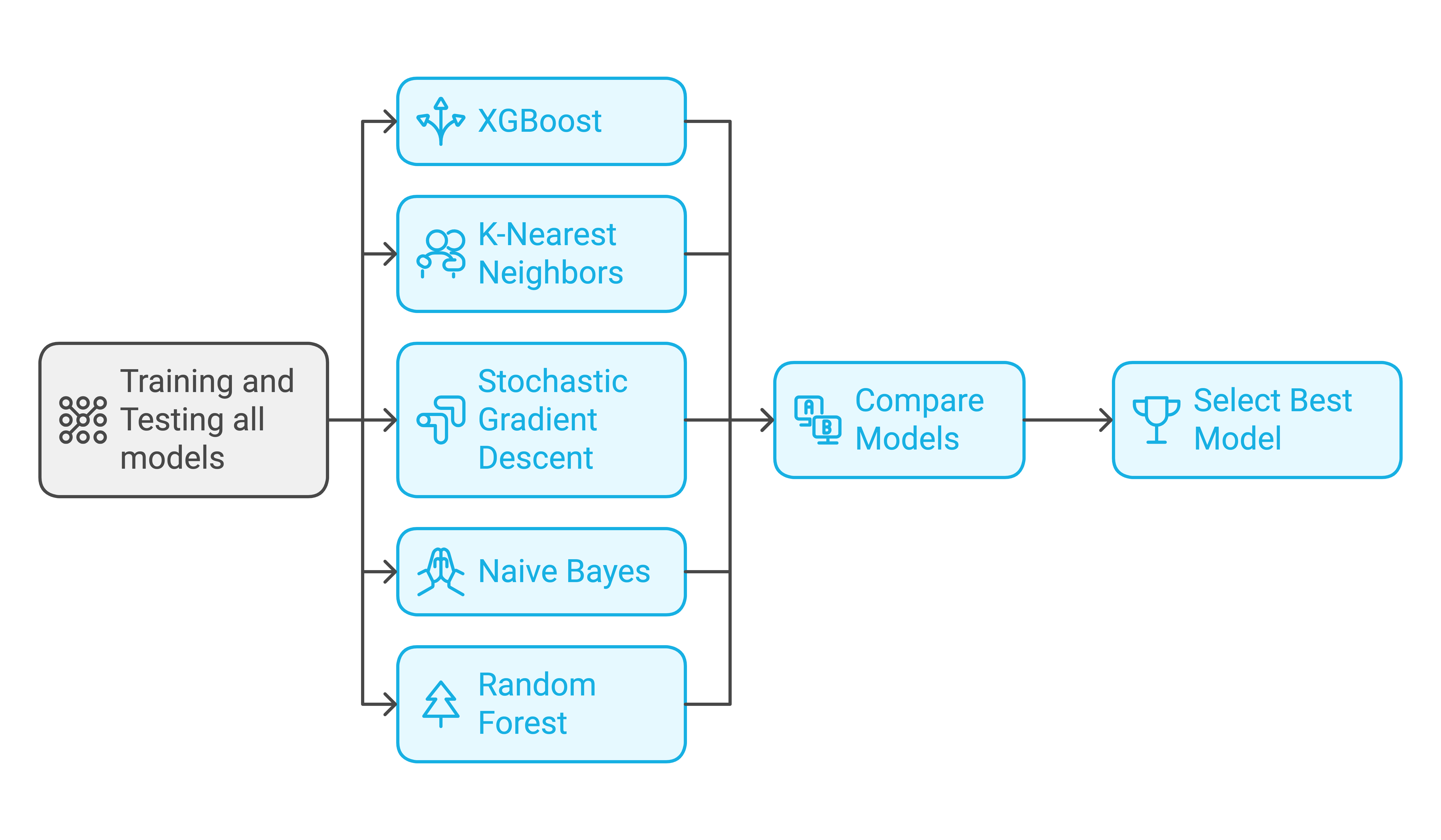}
\caption{Comparison of Machine Learning Models}
\label{fig:model_comparison}
\end{figure}

\subsubsection{\textbf{XGBoost}}
The gradient-boosted decision tree model XGBoost \cite{dhaliwal2018} is meant to be quick and efficient. It works well with complicated datasets because it introduces regularisation to reduce overfitting. The model optimizes an objective function \( \mathcal{L} \), which includes a loss term and a regularization term:
\begin{equation}
    \mathcal{L} = \sum_{i=1}^{n} l(y_i, \hat{y}_i) + \sum_{k=1}^{K} \Omega(f_k)
\end{equation}
where \( l(y_i, \hat{y}_i) \) is the loss function and \( \Omega(f_k) \) represents the regularization applied to each tree.

\vspace{\baselineskip}

\subsubsection{\textbf{RandomForest}}
Random Forest is an ensemble learning methodology whereby many decision trees are constructed during training \cite{dhaliwal2018}. It handles big datasets with grace and also reduces overfitting. A class is chosen based on the collective vote provided by the individual trees of the forest. Typically, decision trees are split at the nodes depending on entropy or Gini Impurity Index:
\begin{equation}
    \text{Gini} = 1 - \sum_{i=1}^{C} p_i^2
\end{equation}
where \( p_i \) is the proportion of instances which belong to class \( i \).

\vspace{\baselineskip}

\subsubsection{\textbf{K-Nearest Neighbors (KNN)}}
KNN is an instance-based learning technique that uses the majority class of its \( k \) closest neighbours to classify a sample. Euclidean distance is used to determine the separation between data points:
\begin{equation}
    d(x, y) = \sqrt{\sum_{i=1}^{n} (x_i - y_i)^2}
\end{equation}
\vspace{\baselineskip}

\subsubsection{\textbf{Stochastic Gradient Descent (SGD)}}
SGD is an optimisation technique that iteratively updates model parameters to minimize a loss function:
\begin{equation}
    \theta := \theta - \eta \nabla_\theta J(\theta)
\end{equation}
where \( \theta \) represents the model parameters, \( \eta \) is the learning rate, and \( J(\theta) \) is the cost function.
\vspace{\baselineskip}

\subsubsection{\textbf{Naïve Bayes}}
Naïve Bayes is a probabilistic classifier that assumes feature independence, using Bayes' theorem:
\begin{equation}
    P(C_k | x) = \frac{P(x | C_k) \cdot P(C_k)}{P(x)}
\end{equation}

\vspace{\baselineskip}

\subsection{Evaluation Metrics}

\begin{figure}[h!]
    \centering
    \includegraphics[width=0.5\textwidth]{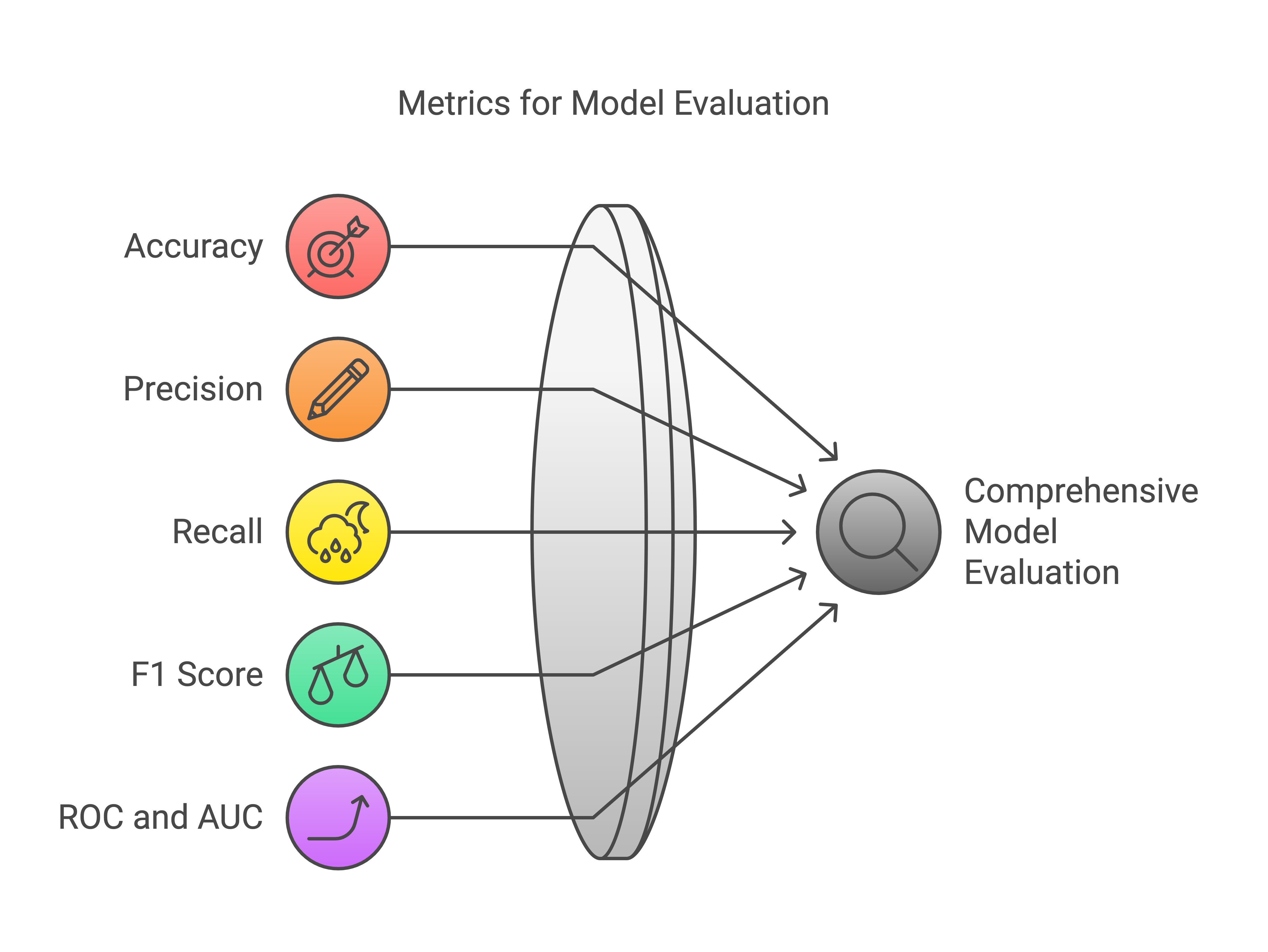}
    \caption{Metrics used for model evaluation}
    \label{fig:model_evaluation_metrics}
\end{figure}

Evaluation Metrices used are as follows:
\begin{itemize}
    \item \textbf{Accuracy}: 
    \begin{equation}
        \text{Accuracy} = \frac{\text{TP + TN}}{\text{TP + TN + FP + FN}}
    \end{equation}
    \item \textbf{Precision}:
    \begin{equation}
        \text{Precision} = \frac{\text{TP}}{\text{TP + FP}}
    \end{equation}
    \item \textbf{Recall}:
    \begin{equation}
        \text{Recall} = \frac{\text{TP}}{\text{TP + FN}}
    \end{equation}
    \item \textbf{F1 Score}:
    \begin{equation}
        \text{F1} = 2 \cdot \frac{\text{Precision} \cdot \text{Recall}}{\text{Precision + Recall}}
    \end{equation}
    \item \textbf{AUC}:
    \begin{equation}
        \text{AUC} = \int_{0}^{1} \text{TPR}(\text{FPR}) \, d(\text{FPR})
    \end{equation}
\end{itemize}
\vspace{\baselineskip}

\section{Proposed Framework}
\label{sec:proposed_framework}

The new proposed framework, an integration of the \textbf{Zero-Trust Zero-Touch AI/ML-Enabled IoT Security Framework}, will ensure the implementation of of end-to-end protection in IoT networks through Zero-Trust principles, Zero-Touch provisioning, and AI/ML-based detection of threats. It covers the most important challenges within the IoT security domain: DDoS mitigation, anomaly detection, and secure onboarding, looking toward 5G/6G environments.
where A signature-based intrusion detection system (IDS) for the Internet of Things (IoT) employs a database of one-day attacks known as attack signatures to identify malicious activities in network traffic.
    \begin{figure}[h!]
    \centering
    \includegraphics[width=0.5\textwidth]{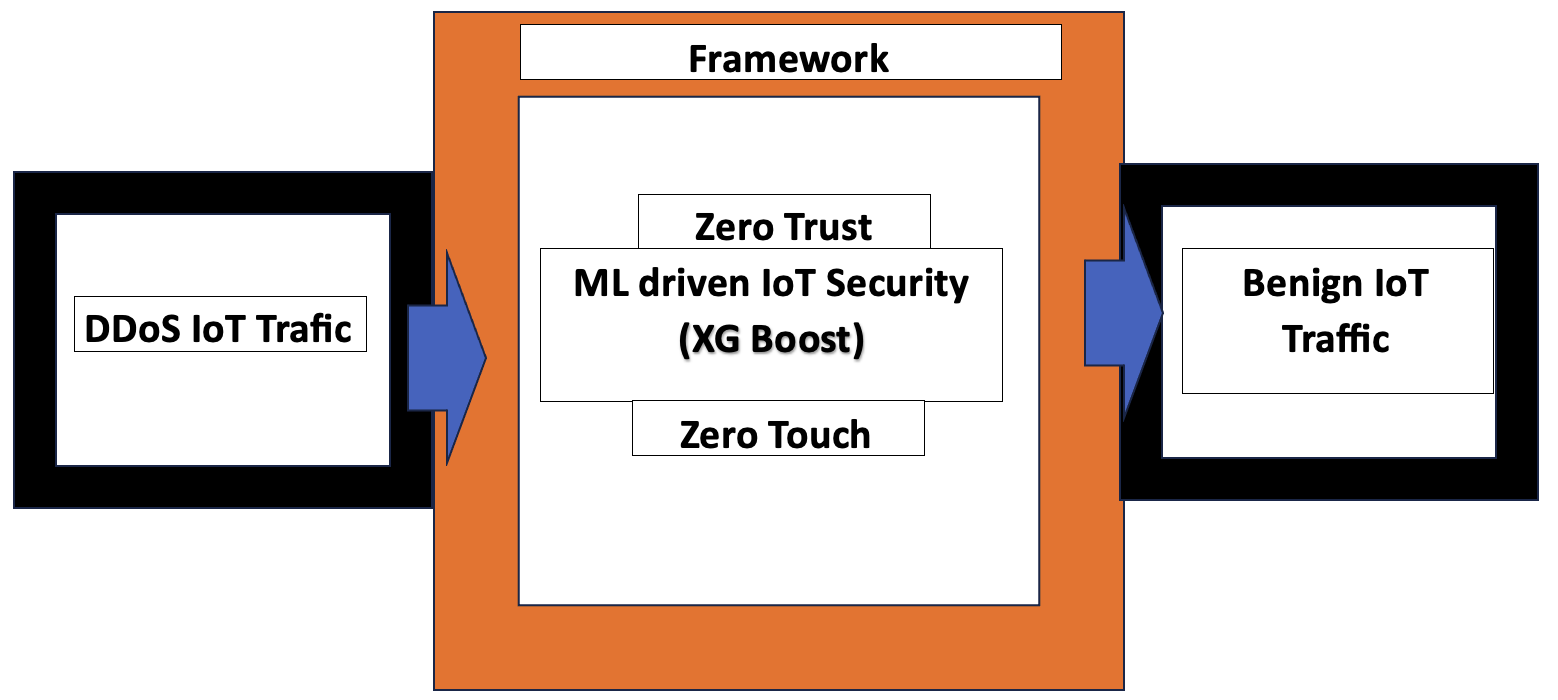}
    \caption{Operational Workflow}
    \label{fig:workflow}
    \end{figure}

\subsection{Framework Architecture}
The framework consists of three core components:
\begin{itemize}
    \item \textbf{Zero-Trust Security}: Ensures that all traffic will be authenticated and verified of all IoT devices, networks, and workloads, and data  and their communications—no more implicit trust, slicing huge margins off the attack surface. The ML-Powered NGFW uses ML-based classifications to intelligently group related IoT devices, depending on network slicing. It can monitor and stop odd and dangerous conduct in this way.\vspace{\baselineskip}
        
    \begin{figure*}[b] 
    \centering \includegraphics[width=1\textwidth]{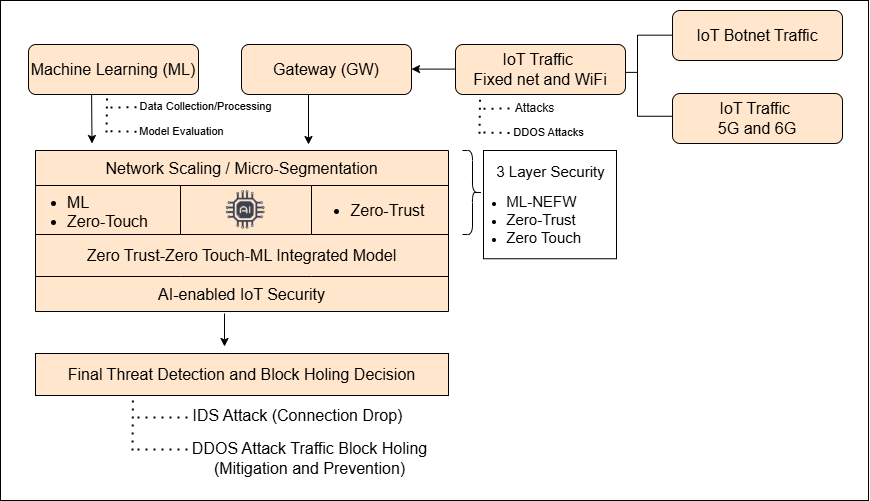} 
    \caption{Zero Trust-Zero Touch Framework} 
    \label{fig:zero_trust_framework} 
    \end{figure*}
    
    \item \textbf{AI/ML-Driven IoT Security}: Advanced machine learning models, including XGBoost, are deployed to enable real-time anomaly detection, DDoS threats, and predictive threat analysis. It would learn and continue to adapt to new threats (zero day)  continuously, providing proactive mitigation against them (one-day threats).
    \vspace{\baselineskip}
    
    \item \textbf{Zero-Touch Automation}: It automates the onboarding of IoT devices, covering secure boot, firmware validation, identity verification, and enabling AI security and network decisions, thus reducing human intervention and configuration errors.\vspace{\baselineskip}
    
\end{itemize}

\subsection{Operational Workflow}
The operational workflow of the framework is illustrated in Fig.~\ref{fig:zero_trust_framework} and involves the following steps:

\begin{enumerate}
    \item \textbf{Input}: IoT traffic coming in is processed, which would include DDoS and benign data.
    \vspace{\baselineskip}
    \item \textbf{Zero-Trust Analysis}: It analyzes the traffic in accordance with Zero-Trust to make sure that any traffic will be transferred only from authenticated or already known devices and communications.

    \vspace{\baselineskip}
    \item \textbf{AI/ML Threat Detection}: Using AI/ML algorithms, it analyzes network traffic and its pattern to recognize patterns of anomalies, DDoS attacks, and vulnerabilities that will pop up in real time.
    \vspace{\baselineskip}
    \item \textbf{Zero-Touch Response}: The effective automated responses are block-holing the malicious traffic, isolation of the compromised devices, and triggering alarms for remediating malicious threats.
    \vspace{\baselineskip}
    \item \textbf{Output}: It allows only benign IoT to pass through the network for safe and unhindered IoT operations.
\end{enumerate}

\subsection{Key Features}
The framework includes the following features:
\begin{itemize}
    \item \textbf{Secure Onboarding}: Auto-provisions and configures IoT devices in a secure way to preserve integrity and minimize human errors.
    \vspace{\baselineskip}
    \item \textbf{Adaptive Threat Detection}: They will self-adapt to the ever-evolving threat landscape with the help of machine learning models to identify zero-day vulnerabilities and new attack vectors.
    \vspace{\baselineskip}
    \item \textbf{Blackholing Traffic}: It shall divert malicious traffic into non-routable sinks to avoid affecting key network resources.
    \vspace{\baselineskip}
    \item \textbf{Scalability in 5G/6G Environments}: Scalability for the 5G/6G next-generation network through solving high device density and diversified use cases made possible due to 5G/6G technologies.

\end{itemize}

\subsection{Illustrative Diagram}

\begin{figure*}[p]
\centering
 \rotatebox{90}{  % Rotates the figure by 90 degrees
    \includegraphics[width=650pt]{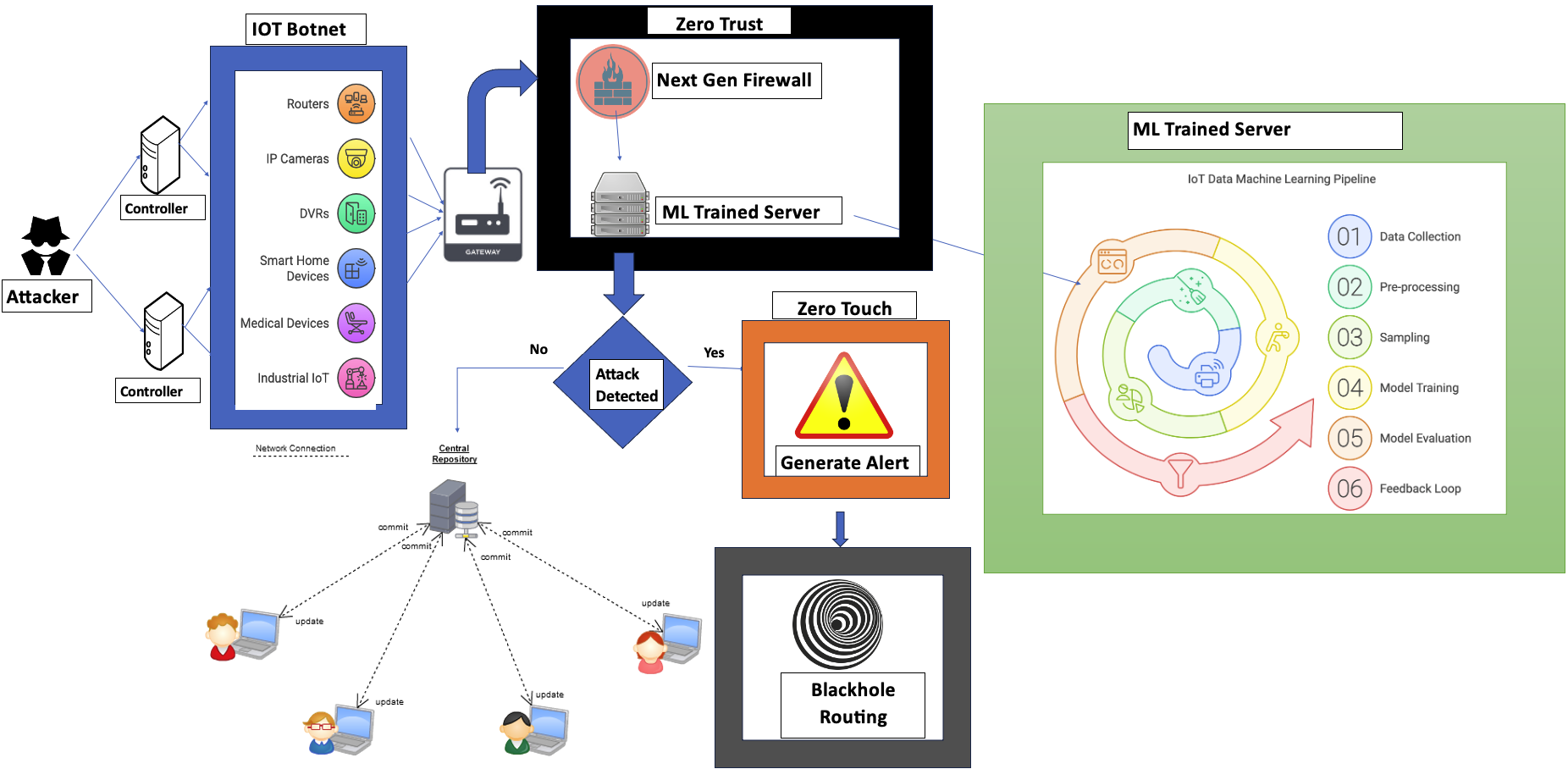}
    }
\caption{Proposed IoT Zero Trust-Zero Touch Framework System Design}
\label{fig:proposed_ztiots}
\end{figure*}

The framework’s architecture and workflow are depicted in Fig.~\ref{fig:proposed_ztiots}, where incoming IoT traffic undergoes a multi-layered Zero-Trust verification process before being granted access to network resources. This approach ensures that no device or user is implicitly trusted, reducing the risk of unauthorized access and lateral movement of threats.

Once traffic is verified, it is analyzed using AI/ML-driven threat detection mechanisms that continuously monitor network behavior for anomalies, suspicious activity, or known attack signatures. These models leverage real-time data streams to identify potential security incidents with high accuracy.

If a threat is detected, the Zero-Touch response system is triggered, automating countermeasures such as isolating compromised devices, blocking malicious traffic, or enforcing adaptive security policies. This automated response ensures rapid threat mitigation without requiring manual intervention, minimizing the impact of cyberattacks on IoT infrastructure.

\subsection{Advantages}
The proposed framework offers several key advantages that enhance IoT security, operational efficiency, and scalability:
\begin{itemize}

    \item \textbf{Proactive Security}: Unlike traditional reactive security models that respond to threats after they occur, this framework adopts a proactive approach by continuously monitoring all incoming IoT traffic through a unified Zero-Trust infrastructure. AI-powered analysis is applied at every checkpoint along the attack surface, allowing the system to detect and mitigate threats before they escalate into larger security incidents. This ensures continuous network operation, preventing downtime and disruption in mission-critical IoT applications such as smart cities, healthcare, industrial IoT, and autonomous systems.
    
    \item \textbf{Operational Efficiency}: The framework automates routine security tasks, such as authentication, anomaly detection, and incident response, reducing the burden on security teams. By minimizing manual intervention, organizations can optimize resource allocation, allowing cybersecurity personnel to focus on higher-priority tasks, such as threat intelligence and policy refinement. This automation-driven security model enhances response times, ensuring rapid containment of potential attacks.
    
    \item \textbf{Improved Scalability}: As IoT ecosystems expand with the advent of 5G and 6G networks, traditional security solutions may struggle to keep up with the increasing number of connected devices. The proposed framework is designed for scalability, enabling seamless integration of new devices without compromising security. It can handle massive volumes of IoT traffic across different network environments, making it well-suited for large-scale smart infrastructure deployments.
    
    \item \textbf{Enhanced Adaptability}: The security framework incorporates continuous learning and AI-driven updates, allowing it to dynamically adapt to emerging threats. As new attack vectors evolve, machine learning models are retrained to recognize novel threats, ensuring that the system remains effective against zero-day vulnerabilities and advanced persistent threats (APTs). This adaptability is particularly critical in IoT environments, where the attack surface is constantly expanding due to the deployment of diverse and heterogeneous devices.
    
\end{itemize}

By integrating Zero-Trust security principles, AI-powered threat detection, and automated response mechanisms, this framework provides a comprehensive and future-ready approach to securing IoT networks. Its ability to proactively identify threats, automate security tasks, and scale with evolving IoT ecosystems makes it a robust solution for mitigating cyber risks in next-generation networks.
\clearpage
\clearpage

\section{Results}

We compare five machine learning models for performance: XGBoost, K-Nearest Neighbors, Stochastic Gradient Descent, Naïve Bayes, and Random Forest. Their metrics are shown in Table~\ref{table:performance}.

\begin{figure}[htbp]
\centerline{\includegraphics[width=\linewidth]{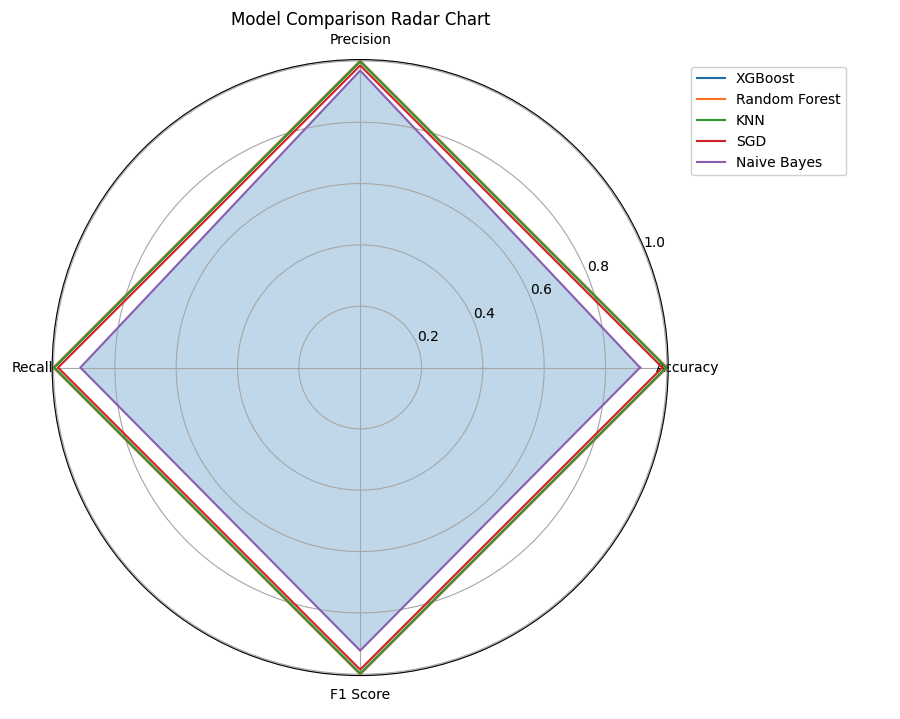}}
\caption{Comparison of Models Performance}
\label{fig:heatmap}
\end{figure}

\subsection{\textbf{XGBoost}}

XGBoost achieved the best overall performance with an accuracy of 99.82\%, precision of 99.82\%, recall of 99.82\%, and an F1 score of 99.82\%, plus the highest AUC of 0.9997. Its built-in regularization effectively mitigates overfitting, making it well-suited to IoT’s dynamic traffic. The gradient-boosting framework iteratively refines weak learners (decision trees), thus excelling at discriminating DDoS traffic from normal activity \cite{pokhrel2021botnet}.

\begin{figure}[h!]
    \centering
    \includegraphics[width=0.5\textwidth]{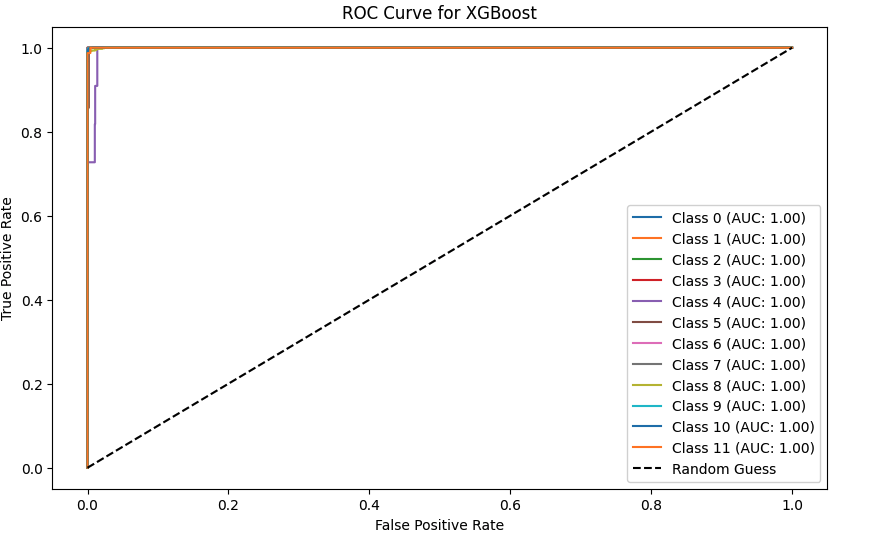}
    \caption{ROC Curve for XGBoost}
    \label{fig:roc_curve_xgboost}
\end{figure}

\subsection{\textbf{Random Forest}}

Random Forest also performed strongly, at 99.79\% for accuracy, precision and recall. In addition, it had a score of 0.9822 for AUC. Although slightly behind XGBoost in AUC, it remains highly robust due to its ensemble of decision trees and inherent feature randomness.

\begin{figure}[ht]
    \centering
    \includegraphics[width=0.5\textwidth]{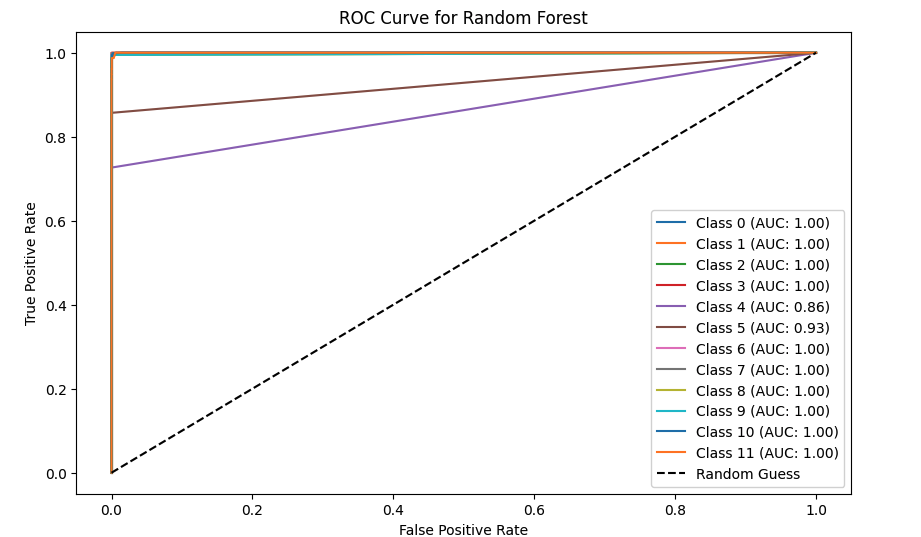}
    \caption{ROC Curve for Random Forest}
    \label{fig:roc_curve_random_forest}
\end{figure}

\subsection{\textbf{K-Nearest Neighbors (KNN)}}
K-Nearest Neighbors (KNN) is a simple yet effective machine learning algorithm that classifies data points based on their proximity to the \textit{k} nearest neighbors in the feature space. In the context of IoT security, KNN achieves an impressive 99.56\%\textbf{ }accuracy, precision, and recall, along with an AUC (Area Under the Curve) of 0.9784, indicating strong classification performance.

The strength of KNN lies in its ease of implementation and interpretability—it does not require an explicit training phase, as classification is performed based on stored data. This makes it particularly useful for anomaly\textbf{ }detection in IoT networks, where distinguishing between normal and malicious behavior is crucial.

However, KNN has scalability\textbf{ }limitations, especially in large-scale IoT networks with massive data streams. Since KNN classifies each new data point by computing the distance to every other point in the dataset, it becomes computationally expensive as the dataset grows. This issue is particularly problematic\textbf{ }in\textbf{ }IoT\textbf{ }environments, where real-time decision-making is often required to detect cyber threats or network anomalies. The high computational burden of distance calculations increases latency, making KNN less suitable for resource-constrained IoT devices.

To mitigate these challenges, optimizations such as KD-trees, Ball trees, or Approximate Nearest Neighbor (ANN) techniques can be employed to accelerate KNN’s performance. Alternatively, other machine learning models such as decision trees, random forests, or deep learning-based approaches may offer a better balance between accuracy and efficiency in large-scale IoT deployments.

\begin{figure}[ht]
    \centering
    \includegraphics[width=0.5\textwidth]{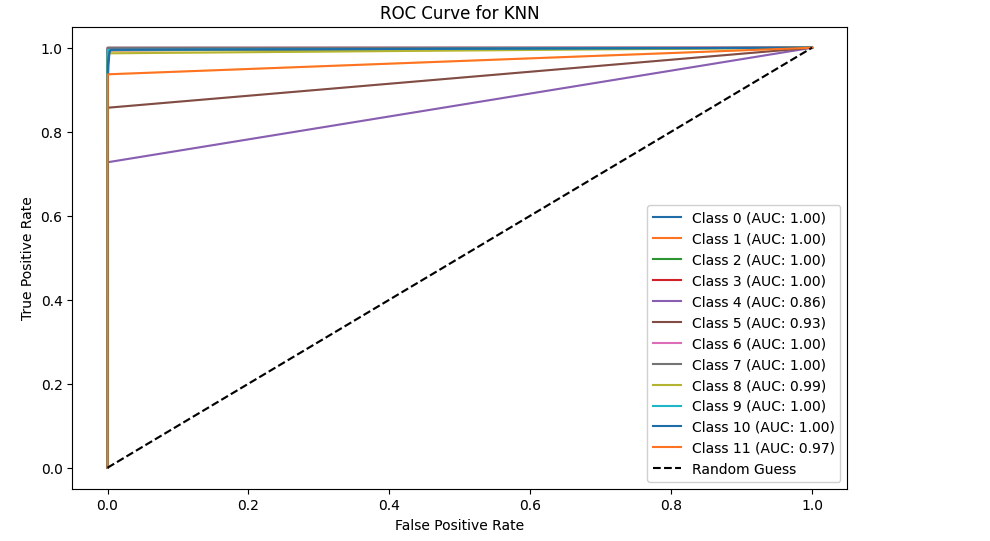}
    \caption{ROC Curve for KNN}
    \label{fig:roc_curve_knn}
\end{figure}

\subsection{\textbf{Stochastic Gradient Descent (SGD)}}

SGD reached 98.52\% accuracy and recall, with an AUC of 0.9868. It is computationally efficient, but can be sensitive to hyperparameters (like learning rate). Although it lags behind the ensemble methods, its speed may be beneficial for real-time edge scenarios \cite{khan2023neural}.

\begin{figure}
    \centering
    \includegraphics[width=0.5\textwidth]{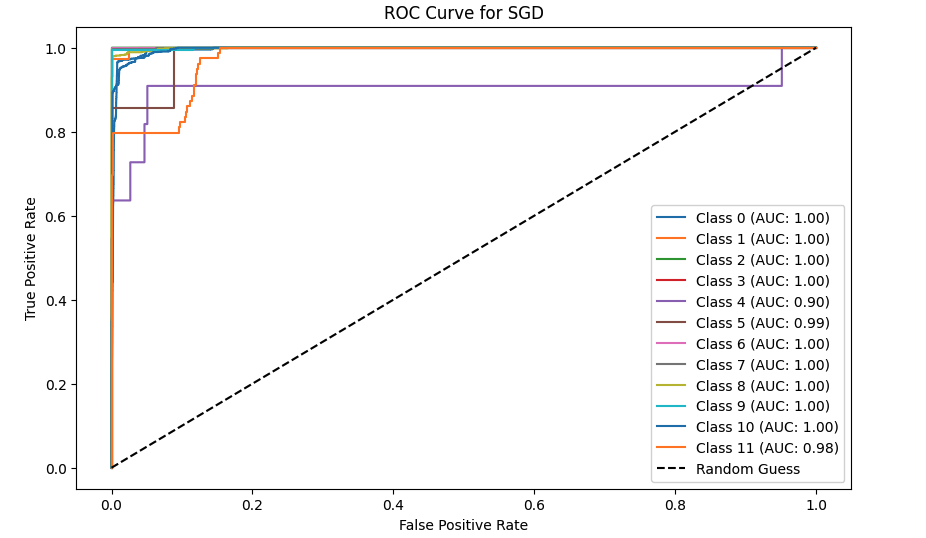}
    \caption{ROC Curve for SGD}
    \label{fig:roc_curve_sgd}
\end{figure}

\subsection{\textbf{Naïve Bayes}}

Naïve Bayes, at 91.09\% accuracy and 0.9829 AUC, suffers from its simplified assumption of feature independence, which limits its capacity to handle complex IoT traffic patterns. Still, its low overhead makes it attractive for ultra-resource-constrained environments.

\begin{figure}
    \centering
    \includegraphics[width=0.5\textwidth]{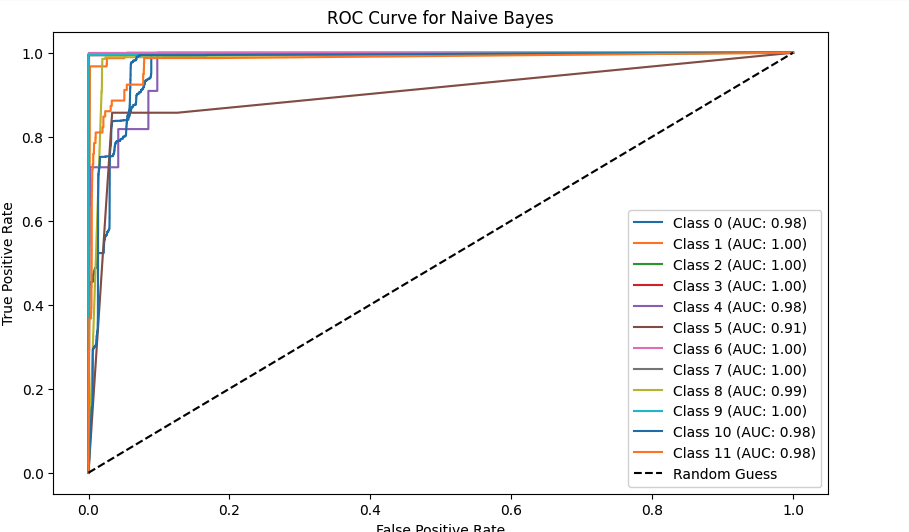}
    \caption{ROC Curve for Naive Bayes}
    \label{fig:roc_curve_naive_bayes}
\end{figure}

\begin{figure}[htbp]
\centerline{\includegraphics[width=\linewidth]{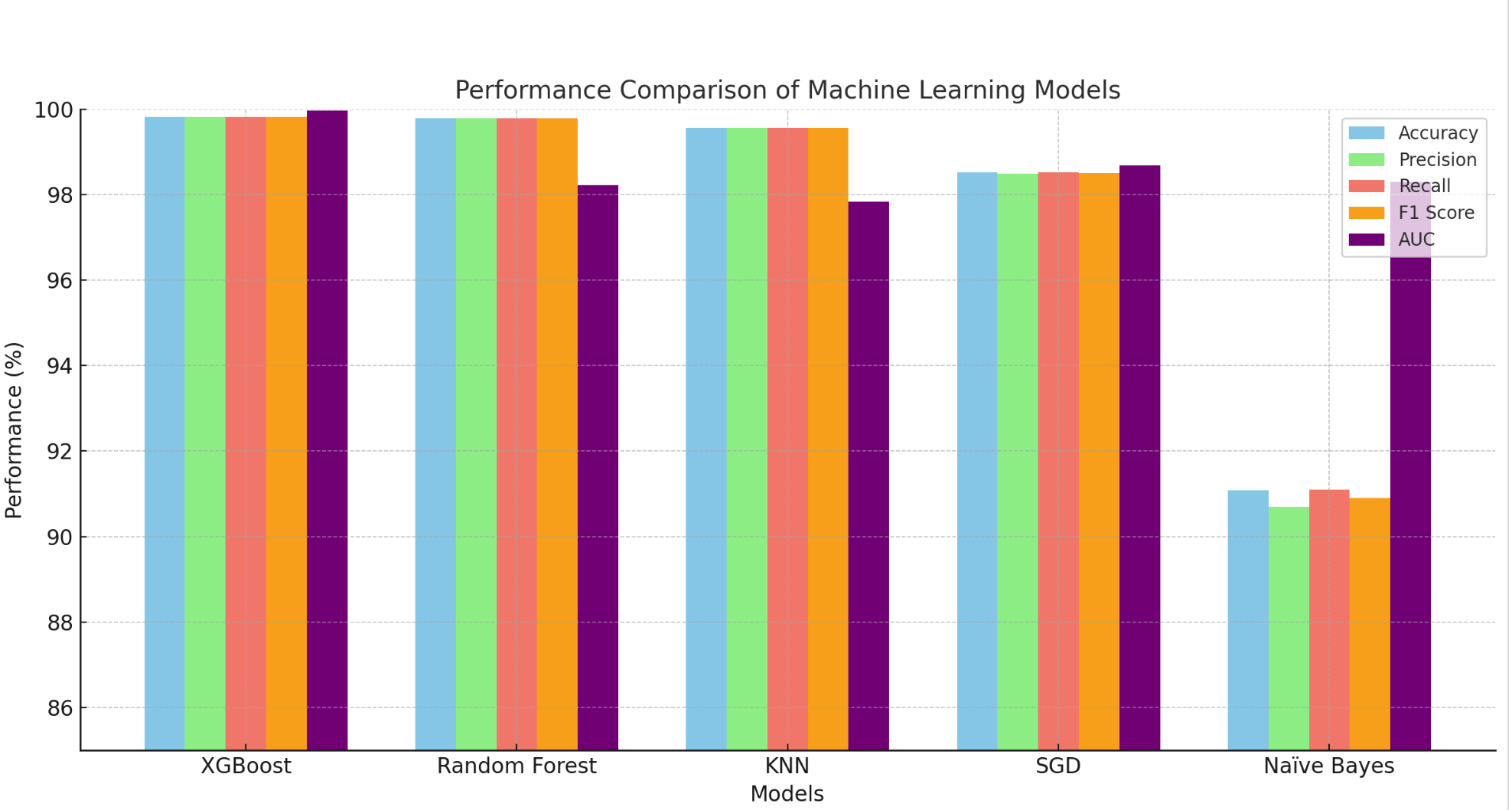}}
\caption{Performance Comparison of ML Models}
\label{fig:results_chart}
\end{figure}

\begin{table*}[!htbp]
\caption{Performance Metrics of Machine Learning Models}
\centering
\resizebox{\textwidth}{!}{%
\begin{tabular}{|c|c|c|c|c|c|}
\hline
\textbf{Model}       & \textbf{Accuracy (\%)} & \textbf{Precision (\%)} & \textbf{Recall (\%)} & \textbf{F1 Score (\%)} & \textbf{AUC (\%)} \\
\hline
XGBoost              & 99.82                 & 99.82                   & 99.82                & 99.82                  & 99.97            \\
Random Forest        & 99.79                 & 99.79                   & 99.79                & 99.79                  & 98.22            \\
KNN                  & 99.56                 & 99.56                   & 99.56                & 99.56                  & 97.84            \\
SGD                  & 98.52                 & 98.49                   & 98.52                & 98.51                  & 98.68            \\
Na\"ive Bayes        & 91.09                 & 90.70                   & 91.10                & 90.90                  & 98.29            \\
\hline
\end{tabular}%
}
\label{table:performance}
\end{table*}

\section{Discussion}
XGBoost emerged as the top performer, excelling in all metrics. Random Forest is nearly as accurate, while KNN, SGD, and Naïve Bayes bring unique advantages in simplicity or speed, albeit with some trade-offs. Ensemble methods (XGBoost, Random Forest) are especially effective at capturing complex patterns in DDoS traffic. This underscores the relevance of advanced ML approaches for real-time IoT security.

The proposed framework demonstrates the viability of integrating Zero-Touch provisioning, Zero-Trust security, and AI/ML-based threat detection for IoT security challenges within 5G/6G environments. Key points from the test results are discussed hereafter:

\begin{itemize}
    \item \textbf{Effectiveness of Ensemble Models}: Ensembling models like XGBoost and Random Forest have been found giving very consistent performance in identifying DDoS attacks with a precision and recall greater than 99 percentage, and the above-discussed capability on complex traffic make them best suited for real-time IoT security.

    \item \textbf{Scalability for Large Networks}: This framework utilizes Zerou Trust to ensure that all traffic is checked and verified. This includes Zero-Touch, which will be utilized for automation for device onboarding and provides security with configuration management, reducing much of the need for manual intervention and therefore also reducing the scope for human error.
  
    \item \textbf{Proactive Threat Mitigation}: The proposed system architecture, through its real-time anomaly detection and automated responses, such as blackholing malicious traffic, will enable proactive threat mitigation much before the threat is capable of taking its toll on IoT operations.

    \item \textbf{Adaptability to Emerging Threats}: It can adapt to emerging threats through continuous learning and identification of new attack vectors with the use of AI/ML-hence (One Day threats) and strong defense mechanisms against zero-day vulnerabilities.
\end{itemize}

\vspace{\baselineskip}

\section{Conclusion}
\label{sec:conclusion}

 The aim of this paper is to propose an integrated framework of \textbf{Zero-Touch provisioning}, \textbf{Zero-Trust security}, and \textbf{AI/ML-based threat detection} for a scalable, adaptive, and proactive approach to the security of modern IoT ecosystems. The framework provides automation of secure device onboarding, continuous authentication, and authorization, and advanced machine learning models for real-time anomaly detection and proactive mitigation of threats. The obtained results ensure the high efficacy of the suggested framework; the results of experiments with ensembles, including XGBoost, show very good accuracy and recall of DDoS attack detection and thus are promising to enable scalability for IoT networks.

The \textbf{Zero-Touch, Zero-Trust, and AI/ML-enabled framework} addresses the modern IoT network for a scalable, efficient and secure solution. This work has ensured that automated provisioning, continuous verification, and adaptive AI-driven threat detection have been guaranteed to ensure resilience against cyber threats in the 5G/6G network era.

\textbf{Key Contributions:}
\begin{itemize}
    \item \textbf{Introduced} a novel integration of Zero-Trust and Zero-Touch principles powered by AI/ML for IoT security.
    \item \textbf{Demonstrated} the effectiveness of ensemble models for detecting and mitigating IoT-based DDoS attacks.
    \item \textbf{Proposed} a scalable framework tailored to the unique challenges of 5G/6G-enabled IoT ecosystems.
    \item \textbf{Provided} automated, proactive security measures that minimize human intervention while ensuring robust network protection.
\end{itemize}

\textbf{Future Directions:} 
While the framework addresses critical IoT security challenges, future work will focus on:
\begin{itemize}
    \item \textbf{Improving} data privacy during training AI/ML models by adopting techniques such as Federated Learning.
    \item \textbf{Reducing} algorithmic bias to make the decisions fair and accurate.
    \item \textbf{Improving} the computational efficiency of ML models for resource-constrained IoT deployments.
    \item \textbf{Making} AI-driven decisions more explainable so that more transparency will be provided for developing trusted automatic security systems.
\end{itemize}

In a nutshell, the proposed framework gives an actionable and inclusive approach toward the security of IoT networks, thus setting the grounds for further research and practical implementation in gradually complex and interrelated environments.

\section*{Acknowledgment}
We gratefully acknowledge all researchers whose pioneering work has shaped IoT security, Zero Trust approach, and ML-based threat detection. Special gratitude is owed to Dr. Robert Abbas for illuminating discussions on our novel framework: zero trust and zero touch security orchestration.

\end{document}